%% file: main.tex
\documentclass[final]{cvpr}
\pdfoutput=1
\usepackage{nopageno}

\usepackage{booktabs} 
\usepackage{times}
\usepackage{animate}
\usepackage{graphicx}
\usepackage{amsmath}
\usepackage{amssymb}
\usepackage[font=small,labelfont=bf,tableposition=top]{caption}
\usepackage[pagebackref=true,breaklinks=true,colorlinks,bookmarks=false]{hyperref}
\usepackage{enumitem}
\input{src/macros} 

\title{SSN: Soft Shadow Network for Image Compositing}
\author{
  Yichen Sheng \\
  Purdue University \\
  \and
  Jianming Zhang \\
  Adobe Research \\
  \and
  Bedrich Benes \\
  Purdue University
}

\begin{document}
\twocolumn[\vspace{-3em}
\maketitle
\input{src/teaser}\bigbreak]
\input{src/0_abstract}


\input{src/1_introduction.tex}
\input{src/2_related_work.tex}
\input{src/3_overview.tex}
\input{src/4_shadows.tex}
\input{src/5_learning.tex}

\input{src/6_results.tex}

\input{src/7_conclusion.tex}

\input{src/8_acknowledgement.tex}

\bibliographystyle{ieee_fullname}
\bibliography{main}
\end{document}

%% file: src/macros.tex
\usepackage{color}
\usepackage{multirow}

\newlength\savewidth\newcommand\shline{\noalign{\global\savewidth\arrayrulewidth\global\arrayrulewidth 1pt}\hline\noalign{\global\arrayrulewidth\savewidth}}

\definecolor{turquoise}{cmyk}{0.65,0,0.1,0.1}
\definecolor{purple}{rgb}{0.65,0,0.65}
\definecolor{darkgreen}{rgb}{0.0, 0.5, 0.0}
\definecolor{darkred}{rgb}{0.5, 0.0, 0.0}
\definecolor{darkblue}{rgb}{0.0, 0.0, 0.5}
\definecolor{blue}{rgb}{0.0, 0.0, 1.0}
\definecolor{orange}{rgb}{1.0,0.5,0.0}

\newcommand{\hide}[1]{{}}

\makeatletter
\renewcommand{\paragraph}{%
  \@startsection{paragraph}{4}%
  {\z@}{0.3ex \@plus 1ex \@minus .1ex}{-1em}%
  {\normalfont\normalsize\bfseries}%
}
\makeatother

%% file: src/teaser.tex
\setlength{\tabcolsep}{0pt}
\centering
\begin{tabular}{lcccc}
    \includegraphics[width=0.19\linewidth]{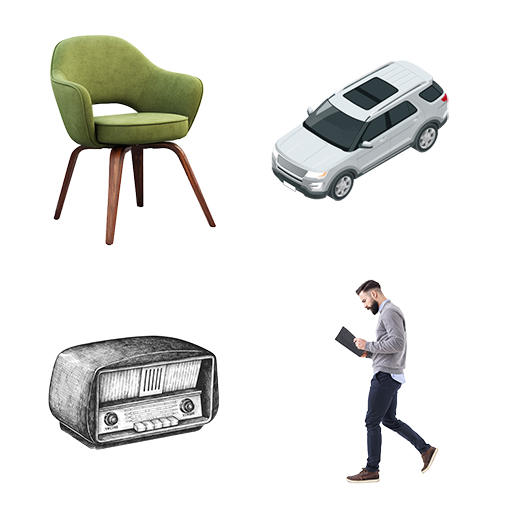} &  
      \animategraphics[autoplay,loop,width=0.19\linewidth]{24}{figs/teaser/chair_pic/p_}{0000}{0127}  & 
      \animategraphics[autoplay,loop,width=0.19\linewidth]{24}{figs/teaser/car_vec/p_}{0000}{0127} &  
      \animategraphics[autoplay,loop,width=0.19\linewidth]{24}{figs/teaser/radio_ske/p_}{0000}{0127} &  
      \animategraphics[autoplay,loop,width=0.19\linewidth]{24}{figs/teaser/man_pic/p_}{0000}{0127}
       \\
   (a) object cutouts (2D masks)&
   \multicolumn{4}{c}{(b) soft shadows generated by our SNN from the image-based light map above}\\
\end{tabular} 

\captionof{figure}{\textbf{Open in Adobe Acrobat to see the animations.} Our Soft Shadow Network (SSN) produces convincing soft shadows given an object cutout mask and a user-specified environment lighting map. (a) shows the object cutouts for this demo, including different object categories and image types, \eg sketch, picture, vector arts. In (b), we show the soft shadow effects generated by our SSN. The changing lighting map used for these examples is shown at the corner of (b). The generated shadows have realistic shade details near the object-ground contact points and enhance image compositing 3D effect. 
}\label{fig:teaser}

%% file: src/0_abstract.tex
\begin{abstract}
We introduce an interactive Soft Shadow Network (SSN) to generates controllable soft shadows for image compositing. 
SSN takes a 2D object mask as input and thus is agnostic to image types such as painting and vector art. An environment light map is used to control the shadow's characteristics, such as angle and softness. SSN employs an Ambient Occlusion Prediction module to predict an intermediate ambient occlusion map, which can be further refined by the user to provides geometric cues to modulate the shadow generation.  To train our model, we design an efficient pipeline to produce diverse soft shadow training data using 3D object models. In addition, we propose an inverse shadow map representation to improve model training. We demonstrate that our model produces realistic soft shadows in real-time. 
Our user studies show that the generated shadows are often indistinguishable from shadows calculated by a physics-based renderer and users can easily use SSN through an interactive application to generate specific shadow effects in minutes. 
\end{abstract}

%% file: src/1_introduction.tex
\begin{figure*}[t]
\setlength{\tabcolsep}{1.0pt}
\begin{tabular}{@{}ccc@{}}
         \includegraphics[width=0.33\linewidth]{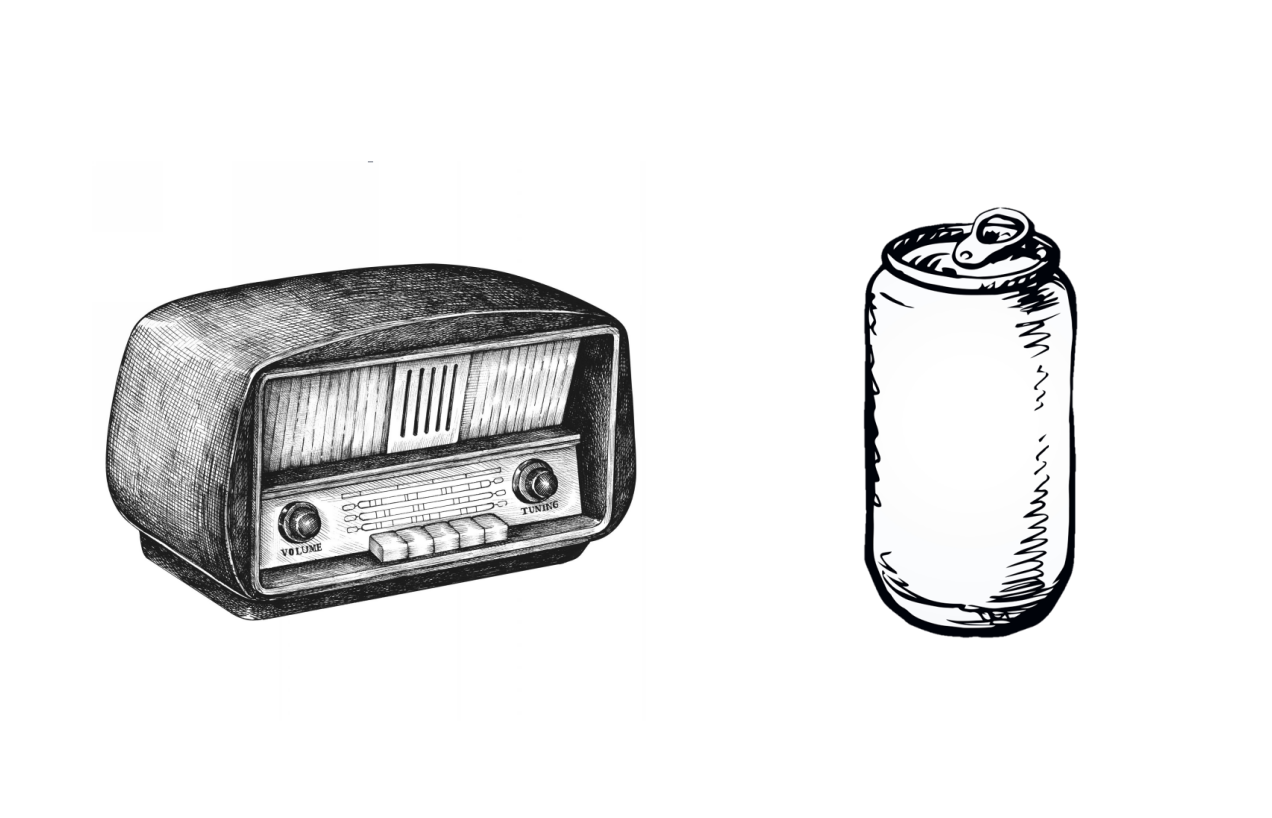} & 
         \includegraphics[width=0.33\linewidth]{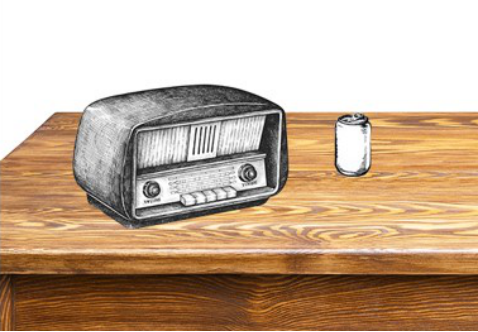}& 
         \includegraphics[width=0.33\linewidth]{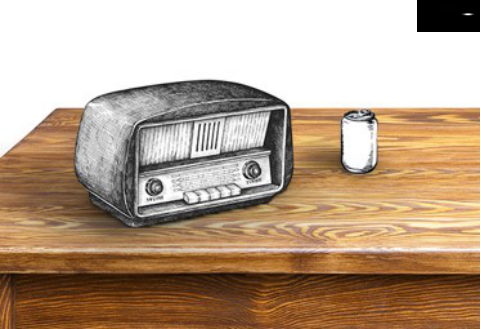}
\end{tabular}
\caption{Image compositing using the Soft Shadow Network (SSN). The user adds two object sketches (left) on a background photo (middle) and uses our SSN to generate realistic soft shadows using a light map shown on the top of the right image. It only takes a couple of minutes for the user to achieve a satisfactory shadow effect. A video records this process can be found in the supplementary material.} \label{fig:comp_example}
\end{figure*}

\section{Introduction}
Image compositing is an essential and powerful means for image creation, where elements from different sources are put together to create a new image. One of the challenging tasks for image compositing is shadow synthesis. Manually creating a convincing shadow for a 2D object cutout requires a significant amount of expertise and effort, because the shadow generation process involves a complex interaction between the object geometry and light sources, especially for area lights and soft shadows. 

Our work eases the creation of soft shadows for 2D object cutouts and provides full controllability to modify the shadow's characteristics. The soft shadow generation requires 3D shape information of the object, which is not available for 2D image compositing. However, the strong 3D shape and pose priors of common objects may provide the essential 3D information for soft shadow generation.

We introduce the Soft Shadow Network (SSN), a deep neural network framework that generates a soft shadow for a 2D object cutout and an input image-based environmental light map.  The input of SSN is an object mask. It is agnostic to image types such as painting, cartoons, or vector arts. User control is provided through the image-based environment light map, which can capture complex light configurations. Fig.~\ref{fig:teaser} show animated shadows predicted by SSN for objects of various shapes in different image types. SSN produces smooth transitions with the changing light maps and realistic shade details on the shadow map, especially near the object-ground contact points.

SSN is composed of an Ambient Occlusion Prediction (AOP) module and a Shadow Rendering (SR) module. Given the object mask, the AOP module predicts an ambient occlusion (AO) map on the ground shadow receiver, which is light map-independent and captures relevant 3D information of the object for soft shadow generation. The SR module then takes the AO map, the object mask, and the light map to generate the soft shadow. Users can refine the predicted AO map when needed to provide extra guidance about the object's shape and region with the ground.

We generate training data for SSN using 3D object models of various shapes and randomly sampled complex light patterns. Rendering soft shadows for complex lighting patterns is time-consuming, which throttles the training process. Therefore, we propose an efficient data pipeline to render complex soft shadows on the fly during the training. In addition, we observe that the shadow map has a high dynamic range, which makes the model training very hard. An inverse shadow map representation is proposed to fix this issue.

A perceptual user study shows that the soft shadows generated by SSN are visually indistinguishable from the soft shadows generated by a physics-based renderer. Moreover, we demonstrate our approach as an interactive tool that allows for real-time shadow manipulation with the system's response to about 5ms in our implementation. As confirmed by a second user study, photo editors can effortlessly incorporate a cutout with desirable soft shadows into an existing image in a couple of minutes by using our tool (see Fig.~\ref{fig:comp_example}). Our main contributions are:
\begin{enumerate}[noitemsep]
\item A novel interactive soft shadow generation framework for generic image compositing.
\item A method to generate diverse training data of soft shadows and environment light maps on the fly.
\item An inverse map representation to improve the training on HDR shadow maps.
\end{enumerate}

%% file: src/2_related_work.tex
\section{Related Work}
\paragraph{Soft Shadow Rendering}
We review soft shadow rendering methods in computer graphics. All these methods require 3D object models, but they are related to our data generation pipeline.

A common method for soft shadow generation from a single area light is its approximation by summing multiple hard shadows~\cite{Crow77}. Various methods were proposed to speed up the soft shadow rendering based on efficient geometrical representations~\cite{agrawala2000efficient, assarsson2003geometry, brabec2002single, EGWR:EGWR03:208-218, guennebaud2006real, HQASSM,schwarz2007bitmask,shadowSil,franke2014delta, cook1984distributed} or image filtering~\cite{Annen08,donnelly2006variance,soler1998fast,heitz2016real}. However, these methods mostly target shadow rendering for a single area light source and are thus less efficient in rendering shadows for complex light settings.

Global illumination algorithms render soft shadows implicitly. Spherical harmonics~\cite{cabral1987bidirectional, sillion1991global, westin1992predicting} based methods render global illumination effects, including soft shadows in real-time by precomputing coefficients in spherical harmonics bases. Instead of projecting visibility function into spherical harmonics bases via expensive Monte-Carlo integration, we use different shadow bases, which are cheaper to compute. 

\paragraph{Image Relighting and shadow Synthesis}
Our method belongs to deep generative models~\cite{goodfellow2014generative, kingma2013auto} performing image synthesis and manipulation via semantic control~\cite{10.1145/3306346.3323023, brock2018large, odena2017conditional} or user guidance such as sketches and painting~\cite{isola2017image,park2019semantic}. 

Deep image harmonization and relighting methods~\cite{shu2017portrait,sun2019single,tsai2017deep,Zheng_2020_CVPR} learn to adapt the subject's appearance to match the target lighting space. This line of works focuses mainly on the harmonization of the subject's appearance, such as color, texture, or lighting style~\cite{sun2019single,sunkavalli2010multi,tao2010error}. 

\begin{figure*}[t]
\begin{center}
    \includegraphics[width=0.9\linewidth]{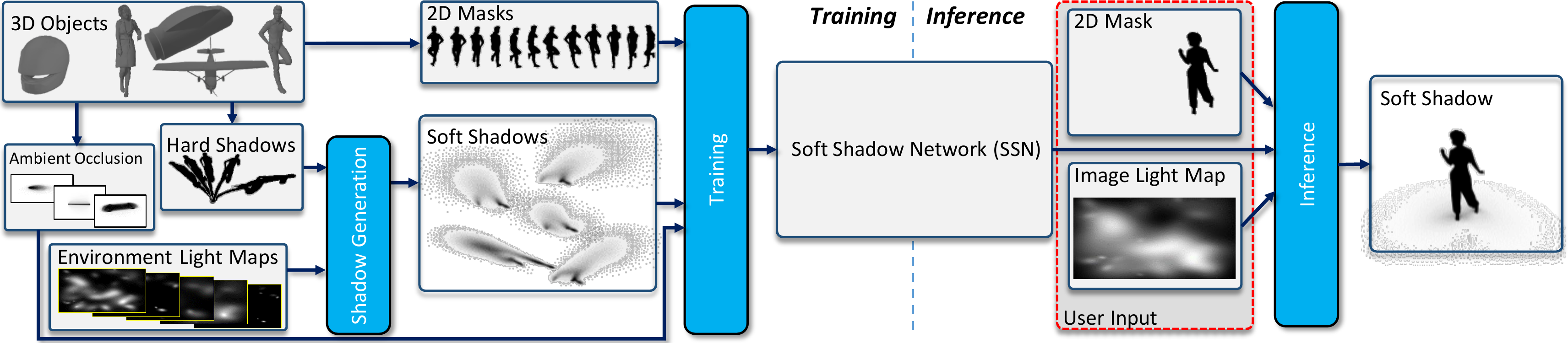}
\end{center}
     \caption{\textbf{System Overview:} During the \textbf{training} phase we train the SSN on a wide variety of 3D objects under different lighting conditions. Each 3D object is viewed from multiple common views, and its 2D mask and hard shadows are computed based on a sampling grid. Hard shadows are processed to become a set of shadow bases for efficient soft shadow computation during training. During the \textbf{inference} step, the user inputs a  2D mask (for example, a cutout from an existing image) and an image light map (either interactively or from a predefined set). The SSN then estimates a soft shadow.}\label{fig:overview}
\end{figure*}

Shadow generation and harmonization can be achieved by estimating the environment lighting and the scene geometry from multiple views~\cite{10.1145/3306346.3323013}. Given a single image, Hold-Geoffroy \etal~\cite{hold2017deep} and Gardner \etal~\cite{gardner2017learning} estimated light maps for 3D object compositing. However, neither the multi-view information nor the object 3D models are available in 2D image compositing. 3D reconstruction methods from a single image~\cite{fan2017point,hassner2006example,knyaz2018image,saito2019pifu} can close this gap. But, they require a complex model architecture design for 3D representation and may not be suitable for time-critical applications such as interactive image editing. Also, the oversimplified camera model in these methods brings artifacts in the touching area between the object and the shadow catcher. Recent works~\cite{hu2019mask,inoue2020rgb2ao,liu2020arshadowgan} explored rendering hard shadows or ambient occlusion using the neural network-based image synthesis methods, but they cannot render soft shadows and lack controls during image editing.

SSN provides a real-time and highly controllable way for interactive soft shadow generation for 2D object cutouts. Our method is trained to infer an object's 3D information for shadow generation implicitly and can be applied in general image compositing tasks for different image types.

%% file: src/3_overview.tex
\section{Overview}

The Soft Shadow Network (SSN) is designed to quickly generate visually plausible soft shadows given a 2D binary mask of a 3D object. The targeted application is image compositing, and the pipeline of our method is shown in Fig.~\ref{fig:overview}. 

The system works in two phases: the first phase trains a deep neural network system to generate soft shadows given 2D binary masks generated from 3D objects and complex image-based light maps. The second phase is the inference step that produces soft shadows for an input 2D binary mask, obtained, for example, as a cutout from an input image. The soft shadow is generated from a user-defined or existing image-based light represented as a 2D image.

The \textbf{training phase} (Fig.~\ref{fig:overview} left) takes as an input a set of 3D objects: we used 186 objects including human and common objects. Each object is viewed from 15 iconic angles, and the generated 2D binary masks are used for training (see Sec.~\ref{sec:3Dobjs}). 

We need to generate soft shadow data for each 3D object. Although we could use a physics-based renderer to generate images of soft shadows, it would be time-consuming. It would require a vast number of soft shadow samples to cover all possible soft shadows combinations with low noise. Therefore, we propose a dynamic soft shadow generation method (Sec.~\ref{sec:shadows}) that only needs to precompute the "cheap" hard shadows once before training. The soft shadow is approximated on-the-fly based on the shadow bases, and the environment light maps (ELMs) randomly generated during the training. To cover a large space of possible lighting conditions, we use Environment Light Maps (ELMs) for lighting. The ELMs are generated procedurally as a combination of 2D Gaussians~\cite{Green06I3D,Yusuke15CGF} (Gaussian mixture) with the varying position, kernel size, and intensity. We randomly sample the ELMs and generate the corresponding soft shadow ground truth in memory on-the-fly during training.
 
The 2D masks and the soft shadows are then used as input to train the SSN as described in Sec.~\ref{sec:SSN}. We use a variant of U-Net~\cite{ronneberger2015u} encoder/decoder network with some additional data injected in the bottleneck part of the network.

The \textbf{inference phase} (Fig.~\ref{fig:overview} right) is aimed at a fast soft shadow generation for image compositing. In a typical scenario, the user selects a part of an image and wants to paste it into an image with soft shadows. The ELM can be either provided or can be painted by a simple GUI. The resulting ELM and the extracted silhouette are then parsed to the SNN that predicts the corresponding soft shadow. 

%% file: src/4_shadows.tex
\section{Dataset Generation}
The input to this step is a set of 3D objects. The output is a set of triplets: a binary masks of the 3D object, an approximated but high-quality soft shadow map of the object cast on a planar surface (floor) from an environment light map (ELM), and an ambient occlusion map for the planar surface. 

\subsection{3D Objects, Masks, and AO Map}\label{sec:3Dobjs}
Let's denote the 3D geometries by $G_i$, where $i=1,\dots,|G|=102$. In our dataset, we used 43 human characters sampled from Daz3D, 59 general objects such as airplanes, bags, bottles, and cars from ShapeNet~\cite{chang2015shapenet} and ModelNet~\cite{wu20153d}. Note that the shadow generation requires only the 3D geometries without textures. Each~$G_i$ is normalized, and its min-max box is put in a canonical position with the center of the min-max box in the origin of the coordinate system. Its projection is aligned with the image's top to fully utilize the image space to receive long shadows.

Each $G_i$ is used to generate fifteen masks denoted by $M_i^j$, where the lower index $i$ denotes the corresponding object $G_i$ and the upper index $j$ is the corresponding view in form $[y,\alpha]$. Each object is rotated five times around the $y$ axis $y=[0^o,45^o,-45^o,90^o,-90^o]$ and is displayed from three common view angles $\alpha=[0^o, 15^o, 30^o]$. This gives the total of $|M_i^j|=1,530$ unique masks (see Fig.~\ref{fig:overview}).
\begin{figure}[t]
\begin{tabular}{cc}
      \includegraphics[width=0.2\linewidth]{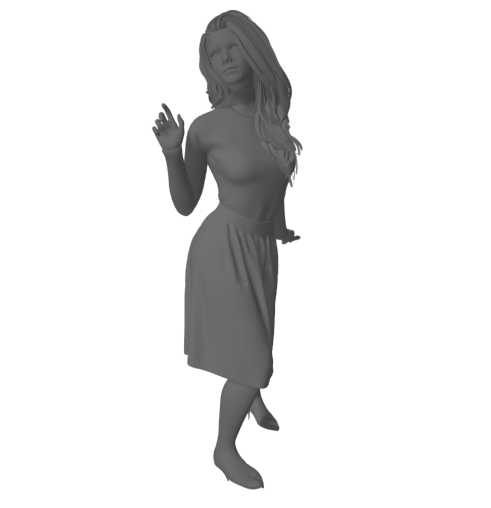} &  
      \includegraphics[width=0.73\linewidth]{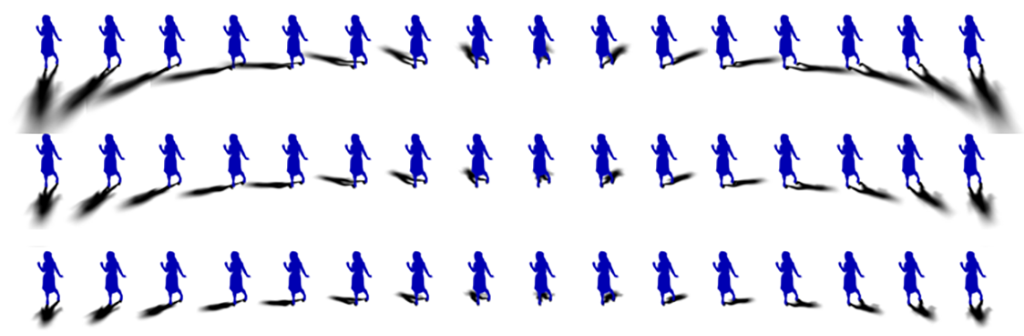}
\end{tabular}
\caption{\textbf{Shadow base example:} for each view of each 3D object, we generate $8\times32$ shadow bases ($3\times16$ shown here). We reduce the soft shadow sampling problem during training to environment light map generation problem, because we use the shadow bases to approximate soft shadows.}\label{fig:geometries}
\end{figure}

Ambient occlusion (AO) map describes how a point is exposed to ambient light (zero: occluded, one: exposed). We calculate the AO map for the shadow receiver (floor) and store it as an image. 
\begin{equation}
    A(x, n) = \frac{1}{\pi} \int_{\Omega} V(x, \omega) \cdot max(n, \omega) d\omega,
\end{equation}
where $V(x, \omega)$ is the visibility term (value is either zero or one) of point $x$ in the solid angle $\omega$. The AO map approximates the proximity of the geometry to the receiver; entirely black pixels are touching the floor. We apply an exponent of one-third on the $A(x,n)$ for a high contrast effect to keep "most" occluded regions strong while weakening those slightly occluded regions. 

\subsection{Environment Light Maps (ELMs)}\label{sec:ELMs}
The second input to the SSN training phase is the soft shadows (see Fig.~\ref{fig:overview}) that are generated from the 3D geometry of~$G_i$ by using environment light maps with HDR image maps in resolution~$512\times{256}$. We use a single light source represented as a 2D Gaussian function:
\begin{equation}
 L_k=Gauss(r,I,\sigma^2),\label{eqn:gauss}
\end{equation} 
where $Gauss$ is a 2D Gaussian function with a radius~$r$, maximum intensity (scaling factor)~$I$, and softness corresponding to~$\sigma^2$. Each ELM is a Gaussian mixture~\cite{Green06I3D,Yusuke15CGF}:
\begin{equation}
 ELM=\sum_{k=1}^{K} L_k([x,y]),\label{eqn:light}
\end{equation} 
where $[x,y]$ is the position of the light source. The coordinates are represented in a normalized range $[0,1]^2$.

\begin{table}[t]
\centering
\caption{Ranges of the ELM parameters. We use random samples from this space during SSN training.}
\vspace{-2mm}
\label{tab:IBL}
\begin{tabular}{l|cc}
\shline
\textbf{meaning} &\textbf{parameter} & \textbf{values}\\
\hline
number of lights &   $K$        & $1,\dots,50$\\
light location   &   $[x,y]$    & $[0,1]^2$\\
light intensity  &   $I$        & $[0,3]$\\
light softness  &   $\sigma^2$ & $[0,0.1]$\\
\shline
\end{tabular}
\end{table}

We provide a wide variety of ELMs that mimic complex natural or human-made lighting configurations so that the SSN can generalize well for arbitrary ELMs. We generate ELMs by random sampling each variable from Eqns~(\ref{eqn:gauss}) and (\ref{eqn:light}) on-the-fly during training. The ranges of each parameter are shown in Table~\ref{tab:IBL}. The overall number of possible lights is vast. 
Note that the ELMs composed of even a small number of lights provide a very high dynamic range of soft shadows.
Please refer to our supplementary materials for samples from the generated data and the comparison to physically-based rendered soft shadows. 


\subsection{Shadow Bases and Soft Shadows}\label{sec:shadows}  
Although we could use a physics-based renderer to generate physically-correct soft shadows, the rendering time for the vast amount of images would be infeasible. Instead, we use a simple method of summing hard shadows generated by a GPU-based renderer by leveraging light's linearity property. Our approach can generate much more diverse soft shadow than a na\"ive sampling several soft shadows from some directions. 

We prepare our shadow bases once during the dataset generation stage. We assume that only the top half regions in the $256\times512$ ELMs cast shadows since the shadow receiver is a plane. For each $16\times16$ non-overlapping patch in the ELM, we sample the hard shadows cast by each pixel included in the patch and sum the group of shadows as a soft shadow base, which is used during training stages.
 
Each model silhouette mask has a set of soft shadow bases. During training, the soft shadow is rendered by weighting the soft shadow bases with the ELM composed of randomly sampled 2D Gaussian mixtures. 

%% file: src/5_learning.tex
\begin{figure*}[t]
\centering
 \includegraphics[width=0.95\linewidth]{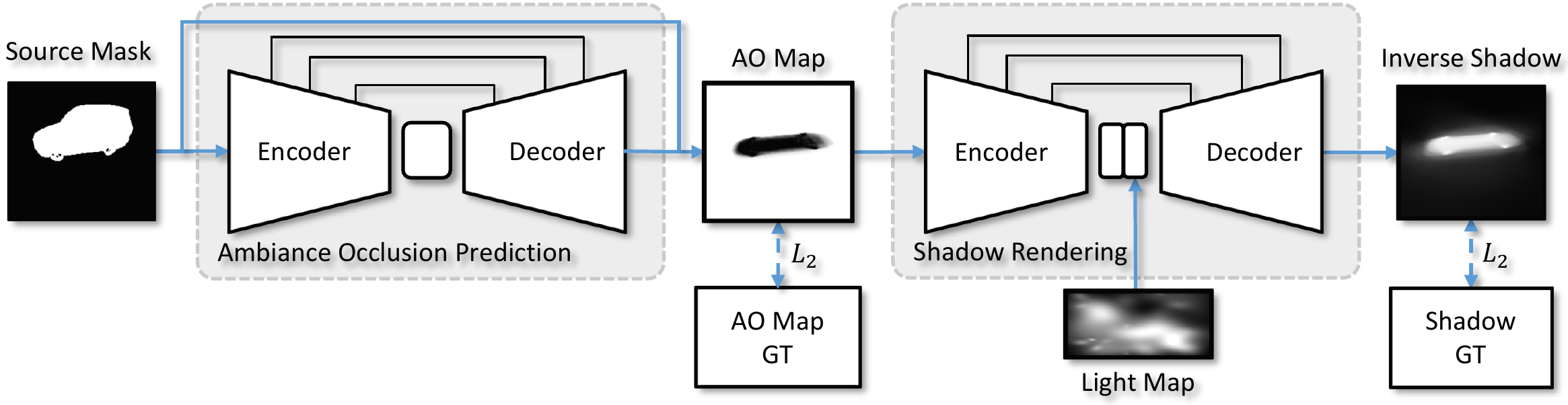}
 \vspace{-7pt}
 \caption{\textbf{SSN Architecture} is composed of two sub-modules: ambient occlusion prediction (AOP) module and shadow rendering (SR) module. AOP module has a U-net shape architecture. It takes a binary mask as input and outputs the ambient occlusion map. SR module has a similar architecture except that we inject the ELM into the bottleneck. Soft shadows are rendered as an output of the SR module. Please refer to supplementary materials for details.}
\label{fig:snn_arch}
\end{figure*}

\section{Learning to Render Soft Shadow}\label{sec:SSN}
We want the model to learn a function $\phi(\cdot)$ that takes cutout mask~$I_m$ and environment light map $I_e$ as input and predicts the soft shadow $I_s$ cast on the ground plane: 
\begin{equation}
    \hat{I}_s = \phi(I_m, L_e).
\end{equation}
During training, the input ELM $L_e$, as described in Sec.~\ref{sec:ELMs}, is generated randomly to ensure the generalization ability of our model. 

During training, we observe that the model has a hard time to converge. This is because the shadow maps have a very wide dynamic range in most of the areas. To address this issue, we propose a simple transform to invert the ground truth shadow map during training:
\begin{equation}\label{shadow_invert}
    \hat{S} = max(S_s) - S_s.
\end{equation}
Therefore, most of the areas except the shadow region is near zero. Inverting the shadows makes the model focus on the final non-zero valued shadow region instead of the high dynamic range of the radiance on the plane receiver. The exact values of the lit radiance are useless for the final shadow prediction. This simple transformation does not bring or lose any information for shadows, but it significantly improves the converging speed and training performance. Please refer to Table~\ref{tab:quantitative_general} for quantitative results.

\subsection{Network Architecture}
SSN architecture (Fig.~\ref{fig:snn_arch}) has two modules: an ambient occlusion prediction (AOP) and a shadow rendering (SR). The overall design of the two modules is inspired by the U-Net~\cite{ronneberger2015u}, except that we inject light source information into the bottleneck of the SR module. In the two modules, both the encoder and decoder are fully convolutional. The AOP module takes masks as input and outputs the ambient occlusion (AO) maps. Then the source masks and the predicted AO maps are passed to the shadow rendering (SR) module. ELM is flattened, repeated in each spatial location, and concatenated with the bottleneck code of the shadow rendering module. The SR module renders soft shadows. The AOP module and the SR module share almost the same layer details. The encoders of both the AOP module and the SR module are composed of a series of $3\times3$ convolution layers. Each convolution layer during encoding follows conv-group norm-ReLU fashion. The decoders applied bilinear upsampling-convolution-group norm-ReLU fashion. We skip link the corresponding activations from corresponding encoder layers to decoder layers for each stage.

\subsection{Loss Function and Training}\label{sec:loss}
The loss for both AOP module and SR module is a per-pixel $L_2$ distance. Let's denote the ground truth of AO map as~$\hat{A}$. The inverse shadow map (Eqn(\ref{shadow_invert})) is~$\hat{S}$, and the prediction of ambient occlusion map as $A$ and soft shadow map as~$S$:
\begin{align}
    L_a(\hat{A}, A) &= ||\hat{A}-A||_2 \label{eqn:ao_loss},\\
    L_s(\hat{S}, S) &= ||\hat{S}-S||_2 \label{eqn:shadow_loss}.
\end{align}

To use big batch size, the AOP and the SR modules are trained separately. For AOP module training, we use the mask as input and compute loss using Eqn~(\ref{eqn:ao_loss}). While for SR module training, we perturb the ground truth AO maps by random erosion and dilation. Then the mask and the perturbed AO map are fed into the SR module for training. During training, we randomly sample the environment light map $ELM$ from Eqn~(\ref{eqn:light}) using our Gaussian mixtures and render the corresponding soft shadow ground truth on the fly to compute shadow loss using Eqn~(\ref{eqn:shadow_loss}). This training routine efficiently helps our model generalize for diverse lighting conditions. The inverse shadow representation also helps the net to converge much faster and leads to better performance. 

We provide a fully automatic pipeline to render soft shadows given a source mask and a target light during the inference stage. We further allow the user to manipulate the target light and modify the predicted ambient occlusion map to better the final rendered soft shadows interactively in real-time.

%% file: src/6_results.tex
\section{Results and Evaluation}

\begin{figure*}[hbt]
\setlength{\tabcolsep}{0.1pt}
\begin{tabular}{ccccccccc}
    \includegraphics[width=0.11\linewidth]{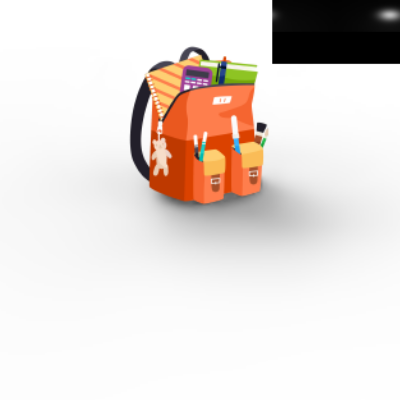}&
    \includegraphics[width=0.11\linewidth]{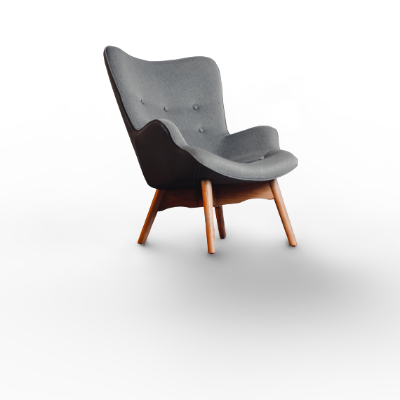}&
    \includegraphics[width=0.11\linewidth]{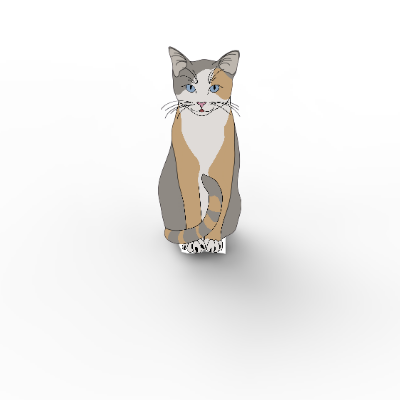}&
    \includegraphics[width=0.11\linewidth]{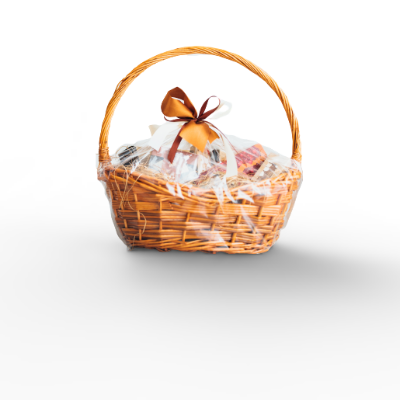}&
    \includegraphics[width=0.11\linewidth]{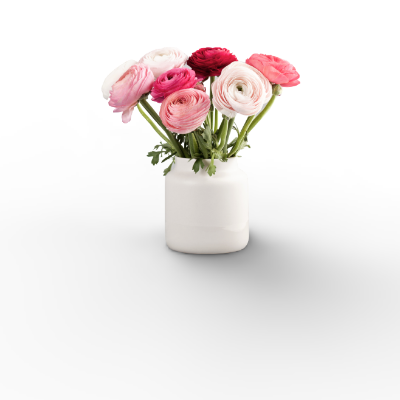}&
    \includegraphics[width=0.11\linewidth]{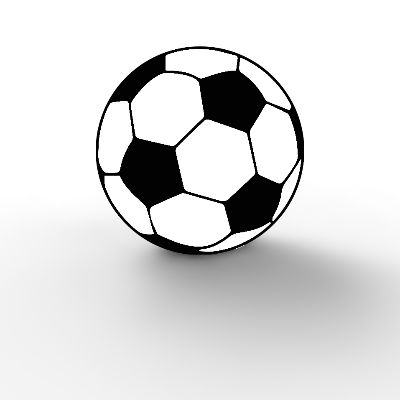}&
    \includegraphics[width=0.11\linewidth]{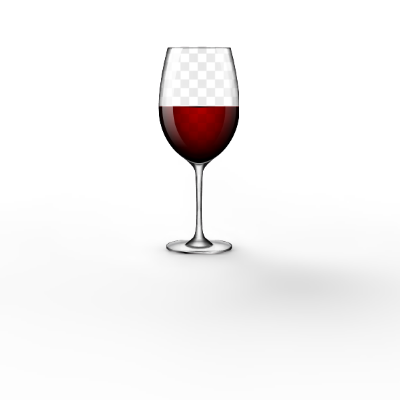}&
    \includegraphics[width=0.11\linewidth]{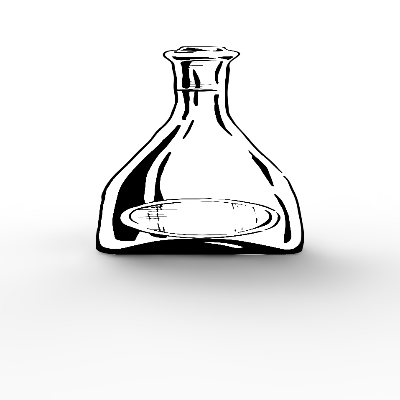}&
    \includegraphics[width=0.11\linewidth]{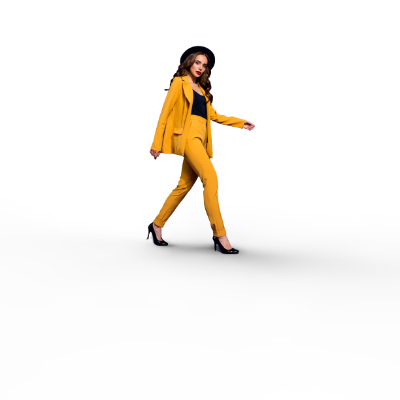}\\
    \includegraphics[width=0.11\linewidth]{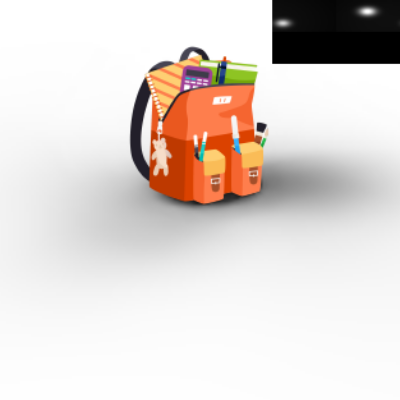}&
    \includegraphics[width=0.11\linewidth]{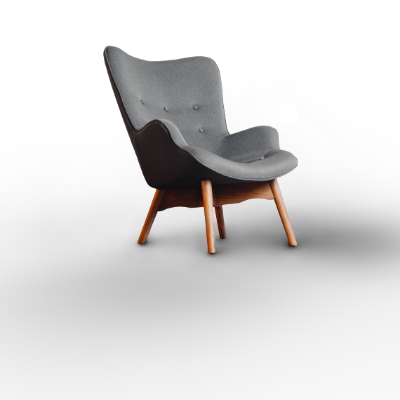}&
    \includegraphics[width=0.11\linewidth]{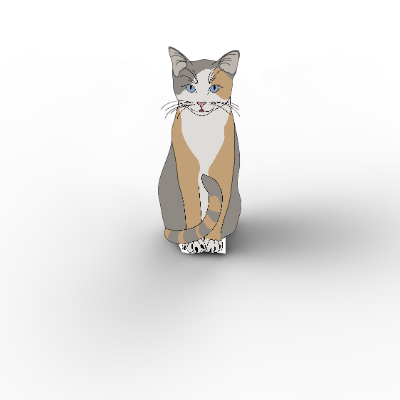}&
    \includegraphics[width=0.11\linewidth]{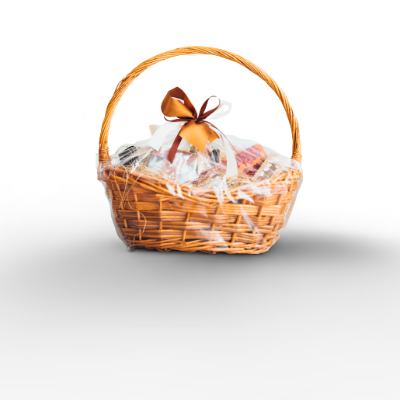}&
    \includegraphics[width=0.11\linewidth]{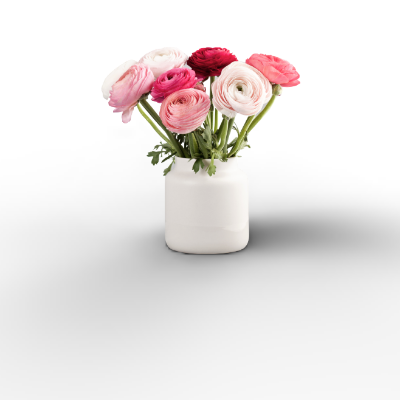}&
    \includegraphics[width=0.11\linewidth]{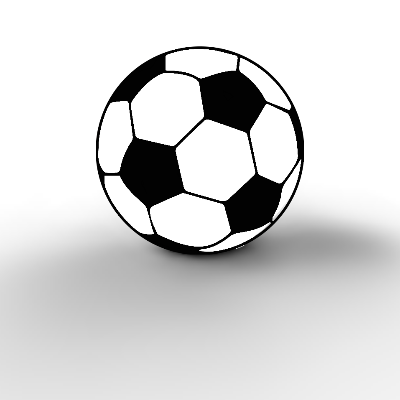}&
    \includegraphics[width=0.11\linewidth]{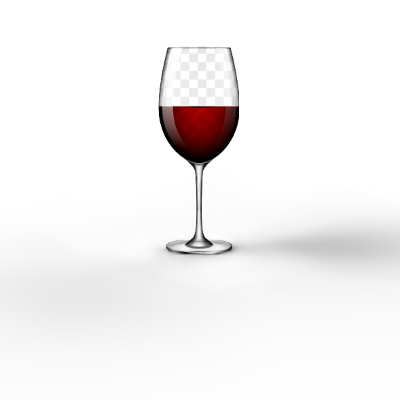}&
    \includegraphics[width=0.11\linewidth]{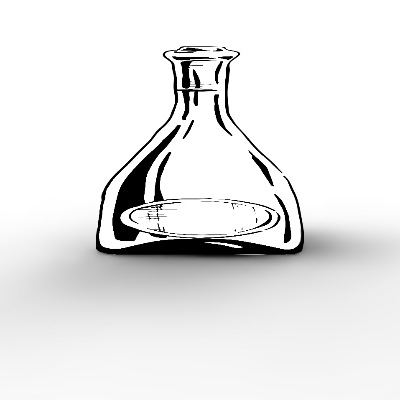}&
    \includegraphics[width=0.11\linewidth]{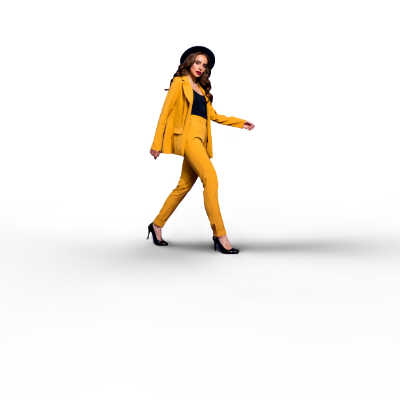}\\
    \includegraphics[width=0.11\linewidth]{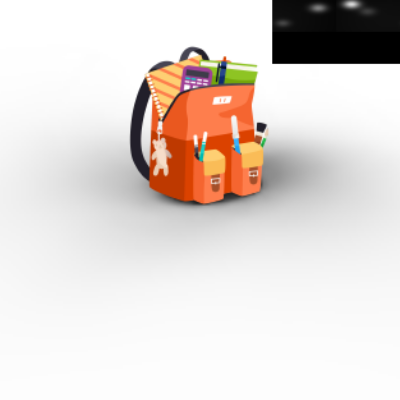}&
    \includegraphics[width=0.11\linewidth]{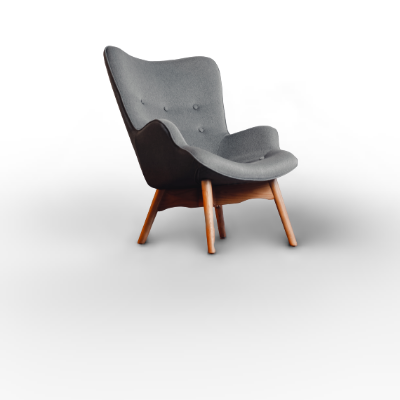}&
    \includegraphics[width=0.11\linewidth]{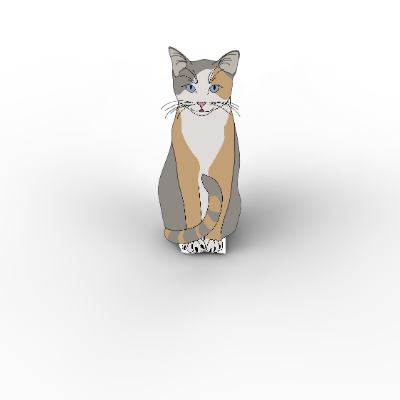}&
    \includegraphics[width=0.11\linewidth]{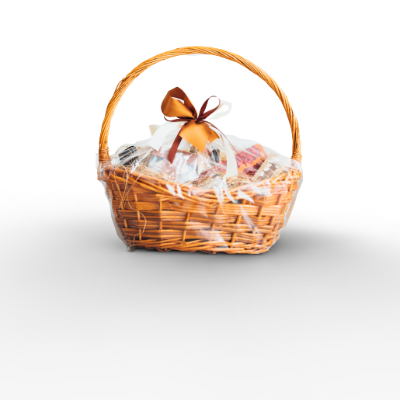}&
    \includegraphics[width=0.11\linewidth]{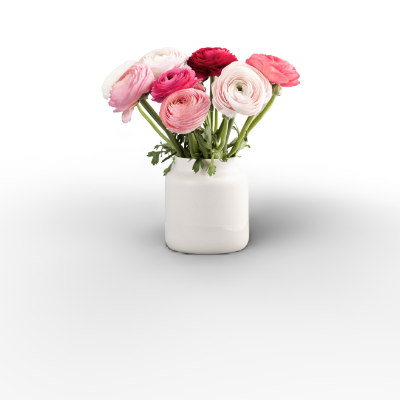}&
    \includegraphics[width=0.11\linewidth]{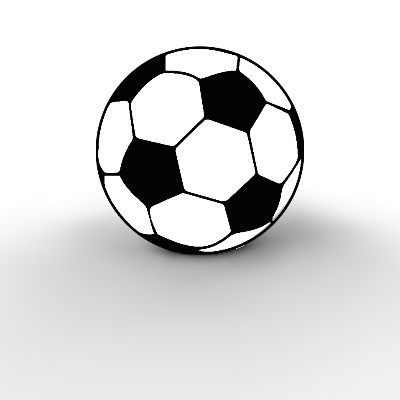}&
    \includegraphics[width=0.11\linewidth]{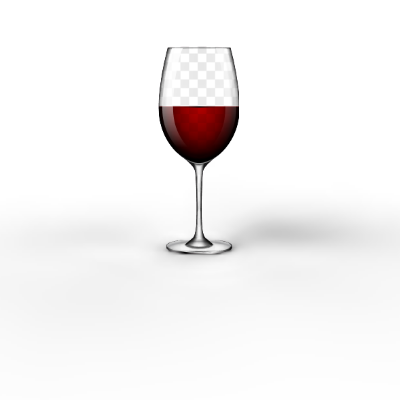}&
    \includegraphics[width=0.11\linewidth]{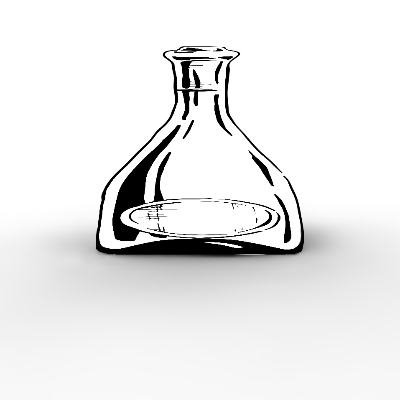}&
    \includegraphics[width=0.11\linewidth]{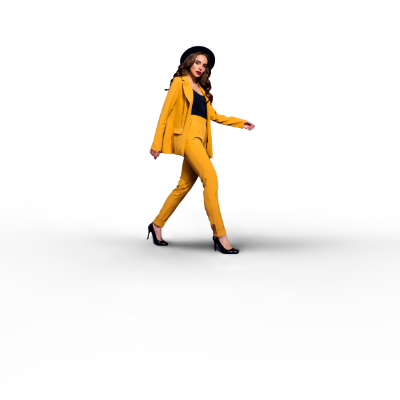}\\
    \includegraphics[width=0.11\linewidth]{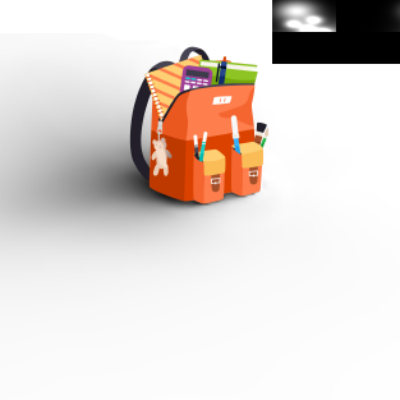}&
    \includegraphics[width=0.11\linewidth]{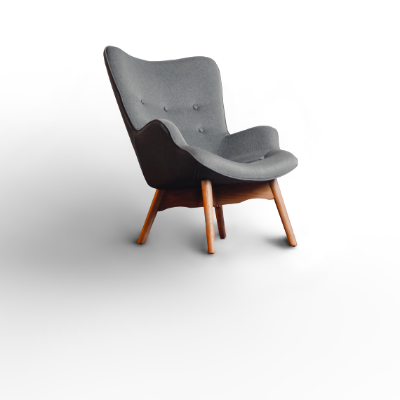}&
    \includegraphics[width=0.11\linewidth]{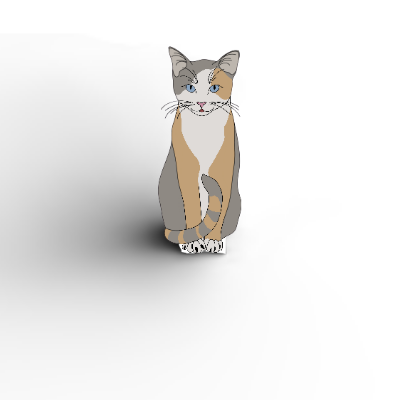}&
    \includegraphics[width=0.11\linewidth]{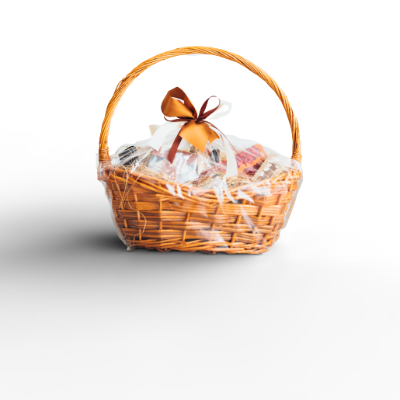}&
    \includegraphics[width=0.11\linewidth]{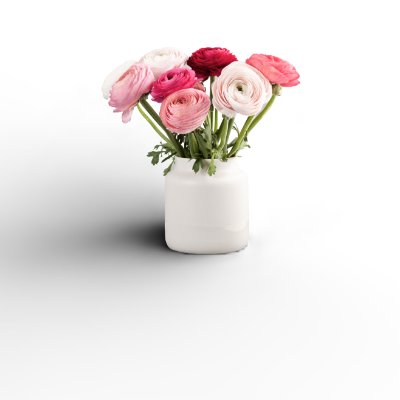}&
    \includegraphics[width=0.11\linewidth]{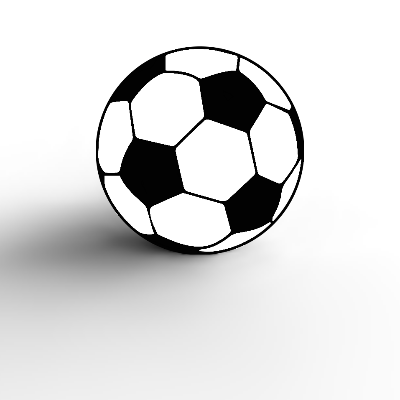}&
    \includegraphics[width=0.11\linewidth]{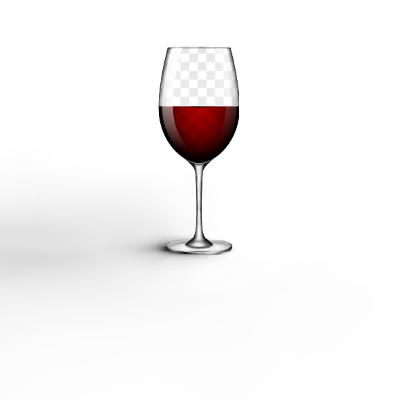}&
    \includegraphics[width=0.11\linewidth]{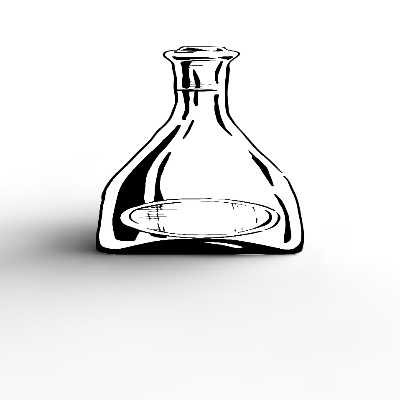}&
    \includegraphics[width=0.11\linewidth]{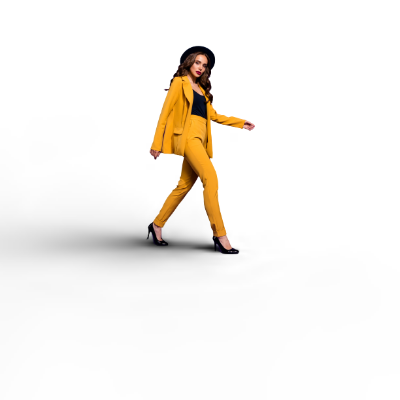}\\
 \end{tabular}
 \caption{Soft shadows generated by SSN using four different light maps. Each row shares the same light map shown in the upper corner of the first image. All light maps have a weak ambient light. The four light maps also have one, two, four, seven strong area lights. Note some objects, \eg cat, football, wine glass, etc, are not covered in our training set.}
 \label{fig:show_case}
\end{figure*}

\subsection{Training Details}
We implemented our deep neural network model by using PyTorch~\cite{paszke2017automatic}. All results were generated on a desktop computer equipped with Intel Xeon W-2145 CPU(3.70GHz), and we used three NVIDIA GeForce GTX TITAN X GPUs for training. We used Adam optimizer~\cite{kingma2013auto} with an initial learning rate of $1e^{-3}$. For each epoch, we run the whole dataset 40$\times$ to sample enough environment maps. Our model converged after 80 epochs and the overall training time was about 40 hours. The average time for soft shadow inference was about $5$ ms.


The animated example in Fig.~\ref{fig:teaser} shows that our method generates smooth shadow transitions for dynamically changing lighting conditions. 

Fig.~\ref{fig:comp_example} shows an example of compositing an existing input scene by inserting several 2D cutouts with rendered soft shadows. Note that once the ELM has been created for one cutout, it is reused by the others. Adding multiple cutouts to an image is simple. 
Several other examples generated by the users are shown in supplementary materials.


\subsection{Quantitative Evaluation}  

\paragraph{Benchmark} 
We separate the benchmark dataset into two different datasets: a general object dataset and a human dataset, 29 other general models from ModelNet~\cite{wu20153d} and ShapeNet~\cite{chang2015shapenet} with different geometries and topologies, \eg~airplane, bag, basket, bottle, car, chair, etc., are covered in the general dataset, nine human models with diverse poses, shapes, and clothes are sampled from Daz3D studio, and we render 15 masks for each model with different camera settings. Soft shadow bases are rendered using the same method in training. We also used the same environment lightmap generation method as described aforementioned to randomly sample 300 different ELMs. Note that all the models are not shown in the training dataset.

\paragraph{Metrics} 
We used four metrics to evaluate the testing performance of SSN: 1)~RMSE, 2)~RMSE-s~\cite{10.1145/3306346.3323008}, 3)~zero-normalized cross-correlation (ZNCC), and 4)~structural dissimilarity (DSSIM)~\cite{schieber2017quantification}. Since the exposure condition of the rendered image may vary due to different rendering implementations, we use scale-invariant metric RMSE-s, ZNCC, and DSSIM in addition to RMSE. Note that all the measurements are computed in the inverse shadow domain.

\paragraph{Ablation study} 
To evaluate the effectiveness of inverse shadow representation and AO map input, we perform ablation study on the aforementioned benchmark and evaluate results by the four metrics. \emph{Non-inv-SSN} denotes a baseline that is the same as SSN but uses the non-inverse shadow representation. \emph{SSN} is our method with inverse shadow representation. \emph{GT-AO-SSN} is a baseline for replacing the predicted AO map with the ground truth one for the SR module. This is to show the upper-bound of improvement by refining the AO map when the geometry of the object is ambiguous to SSN.

Table~\ref{tab:quantitative_general} shows that Non-inv-SSN has a significantly worse performance for each metric. By comparing the metric difference between SSN and GT-AO-SSN in Table~\ref{tab:quantitative_general} and Table~\ref{tab:quantitative_human}, it is observed that in some specific dataset, \eg~human dataset, our SSN has a reasonably good performance without refining the ambient occlusion map. We further validate it by a user study discussed in qualitative evaluation. 
In a more diverse test dataset, Table~\ref{tab:quantitative_general} shows that soft shadow quality is improved with better AO map input. Also, Fig.~\ref{fig:show_case} shows that SSN generalizes well for various unseen objects with different ELMs.

\begin{table}[t]
\centering
\caption{Quantitative shadow analysis on general object benchmark. Non-inv-SSN uses the same architecture with our SSN except that the training shadow ground truth is not inverted. GT-AO-SSN uses ground truth ambient occlusion map as input for SR module. For RMSE, RMSE-s, DSSIM, the lower the values, the better the shadow prediction, while ZNCC is on the opposite.}\label{tab:rms}
\vspace{-2mm}
\small
\begin{tabular}{l|cccc}
\shline
\textbf{Method} &\textbf{RMSE}  & \textbf{RMSE-s}   & \textbf{ZNCC}     & \textbf{DSSIM} \\\hline
Non-inv-SSN     &  0.0926     &   0.0894        & 0.7521          & 0.2913 \\
SSN             &  0.0561     &   0.0506        & 0.8192          & 0.0616 \\
GT-AO-SSN       &  0.0342     &   0.0304        & 0.9171          & 0.0461  \\\shline
\end{tabular}
\label{tab:quantitative_general}
\end{table}

\begin{table}[t]
\centering
\caption{Quantitative shadow analysis on human benchmark. The difference between the two methods is much smaller than the same methods in Table~\ref{tab:quantitative_general}, indicating SSN can have a good performance for some specific object.}
\vspace{-2mm}
\label{tab:quantitative_human}
\small
\begin{tabular}{l|cccc}
\shline
\textbf{Method} &\textbf{RMSE}  & \textbf{RMSE-s} & \textbf{ZNCC}   & \textbf{DSSIM} \\\hline
SSN     &  0.0194    &   0.0163    & 0.8943          & 0.0467 \\
GT-AO-SSN  &  0.0150    &   0.0127    & 0.9316          & 0.0403 \\\shline
\end{tabular}
\end{table}

\subsection{Qualitative Evaluation}
We performed two perceptual user studies. The first one measures the perceived level of realism of shadows generated by the SSN; the second one tested the shadow generator's ease-of-use. 

\begin{figure}[t]
\centering
\includegraphics[width=0.49\linewidth, trim=0 2cm 0 0]{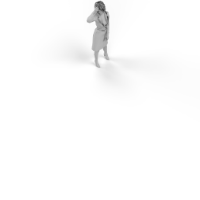}
\includegraphics[width=0.49\linewidth, trim=0 2cm 0 0]{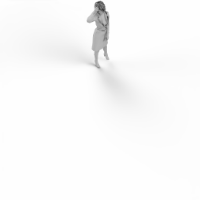}
\caption{A sample pair of images from our Perceived  Realism user study. We show the output generated from 3D objects rendered by Mitsuba (left) and the output generated by SSN from binary masks (right).} \label{fig:2afc}
\end{figure}

\paragraph{Perceived Realism (user study 1)} We have generated two sets of images with soft shadows. One set, called MTR, was generated from the 3D object by rendering it in Mitsuba renderer, a physics-based rendered and was considered the ground truth when using enough samples. The second set, called SSN, used binary masks from the same objects from MTR and estimated the soft shadows. Both sets have the same number of images $|MTR|=|SSN|=18$, resulting in 18 pairs. In both cases, we used 3D objects that were not present in the training set or the SSN validation set during its training. The presented objects were unknown to the SSN. The ELM used were designed to cover a wide variety of shadows ranging from a single shadow, two shadows to a very subtle shape and intensity. Fig.~\ref{fig:2afc} shows an example of the pair of images used in our study. Please refer to supplementary materials for some other user study examples.

\begin{figure}[t]
 \includegraphics[width=0.9\linewidth]{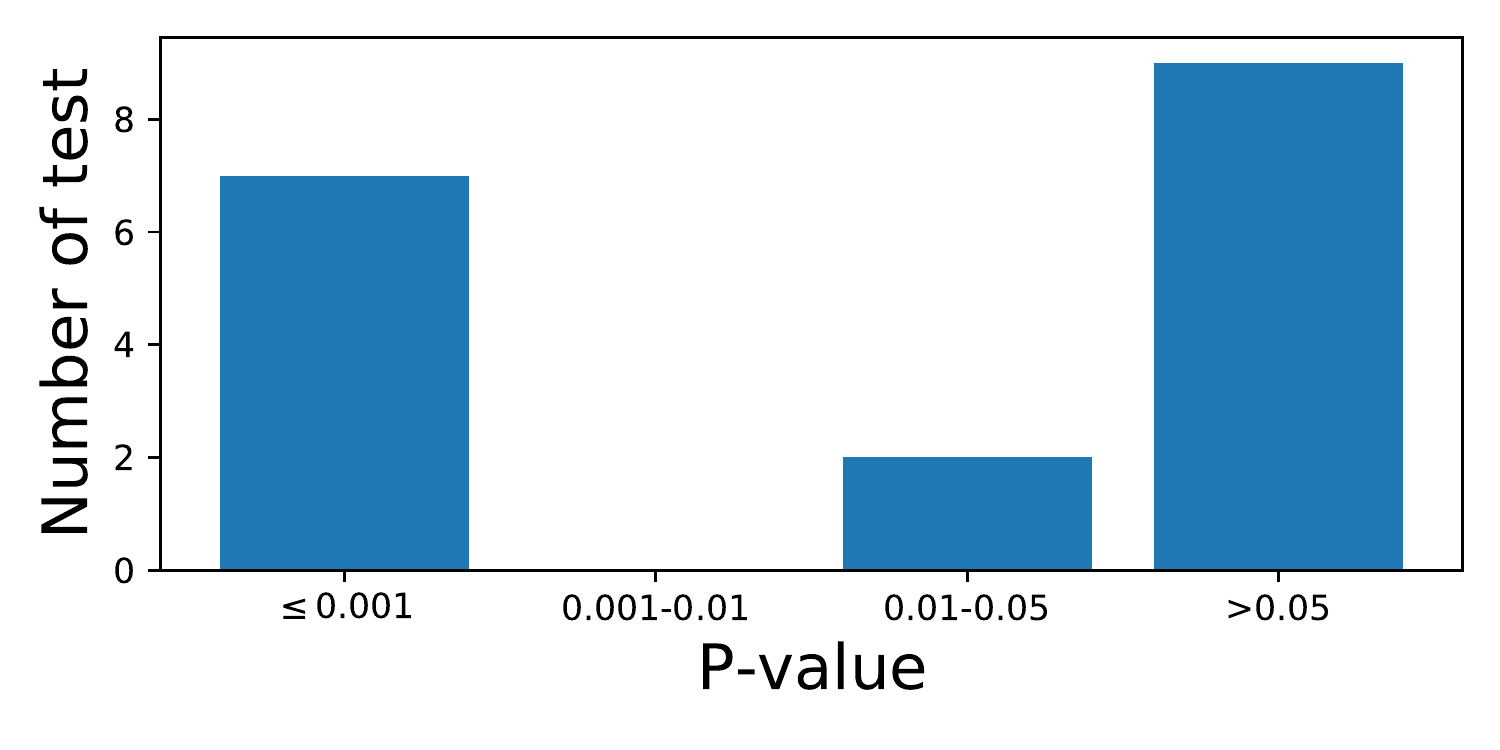}
 \caption{\textbf{p-value distributions:} In our first user study 7 questions have p-value$\leq{0.001}$, two questions  $0.01<\text{p-value}\leq{0.05}$, and 9 questions have $\text{p-value}>0.05$.}
 \label{fig:p-value}
\end{figure}

\begin{figure*}[t]
\setlength{\tabcolsep}{1.0pt}
\centering
\small
\begin{tabular}{cccc}
    \includegraphics[width=0.24\linewidth]{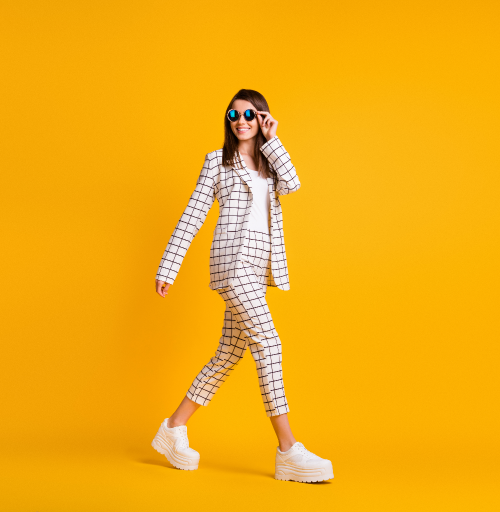}&
    \includegraphics[width=0.24\linewidth]{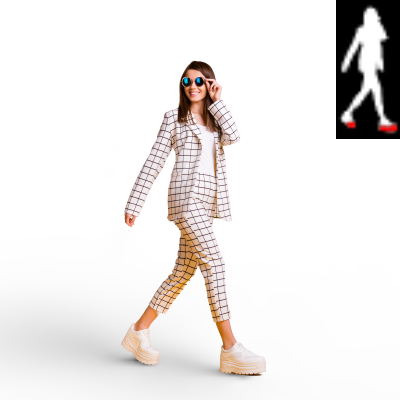}&
    \includegraphics[width=0.24\linewidth]{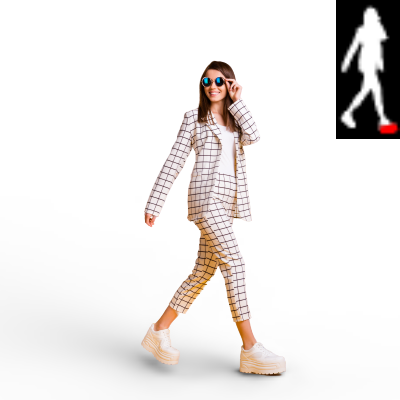}&
    \includegraphics[width=0.24\linewidth]{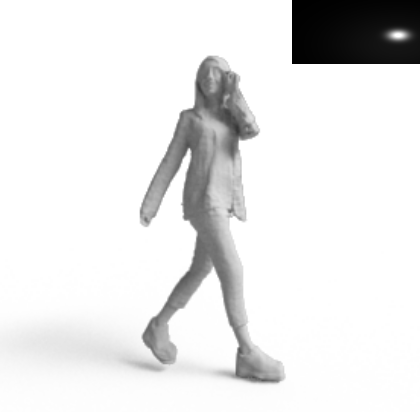} \\
    Input Cutout & SSN & SSN w.~AO Map Modification & PIFuHD + Mitsuba Rendering
\end{tabular}
\caption{Shadow generation comparison with PIFuHD-based approach. SSN (2$^\textrm{nd}$) renders the soft shadow given a cutout from an image (1$^\textrm{st}$). The intermediate AO map prediction from our AO prediction module is visualized by the red regions overlaying the cutout mask shown in the top-right corner. SSN can also render a different soft shadow using a different AO input to change the 3D relationship between the occluder and shadow receiver (3$^\textrm{rd}$). Mitsuba (4$^\textrm{th}$) renders a soft shadow from the reconstructed 3D geometry of the object by PIFuHD, but it is hard to adjust the foot contact to match the original image. The ELM used for the three examples is in the corner of the 4$^\textrm{th}$ image.} \label{fig:comparison_pifuhd}
\end{figure*}

The perceptual study was a two-alternative forced-choice (2AFC) method. To validate the rendered images' perceived realism, we have shown pairs of images in random order and random position (left-right) to multiple users and asked the participants of an online study which of the two images is a fake shadow.

The study was answered by 56 participants (73\% male, 25\% female, 2\% did not identify). We discarded all replies that were too short (under three minutes) or did not complete all the questions. We also discarded answers of users who always clicked on the same side. Each image pair was viewed by 46 valid users. In general, the users were not able to distinguish SSN-generated shadows from the ground truth. In particular, the result shows that the average accuracy was 48.1\% with a standard deviation of 0.153. T-test for each question in Fig.~\ref{fig:p-value} shows that there are half of the predictions that do not have a significant difference with the Mitsuba ground truth.  

\paragraph{Ease-of-Use (user study 2):} In the second study, human subjects were asked to recreate soft shadows by using a simple interactive application. The result shows users can generate specified soft shadows in minutes using our GUI. Refer to supplementary materials for more details.


\begin{figure}[ht]
\centering
\includegraphics[width=0.49\linewidth]{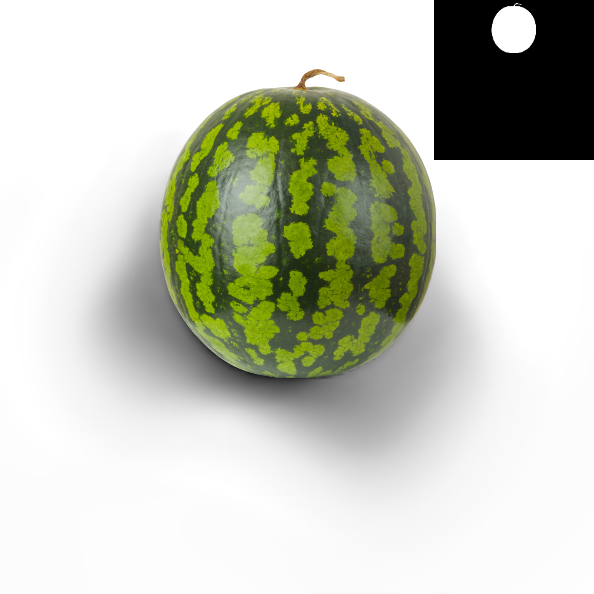}
\includegraphics[width=0.49\linewidth]{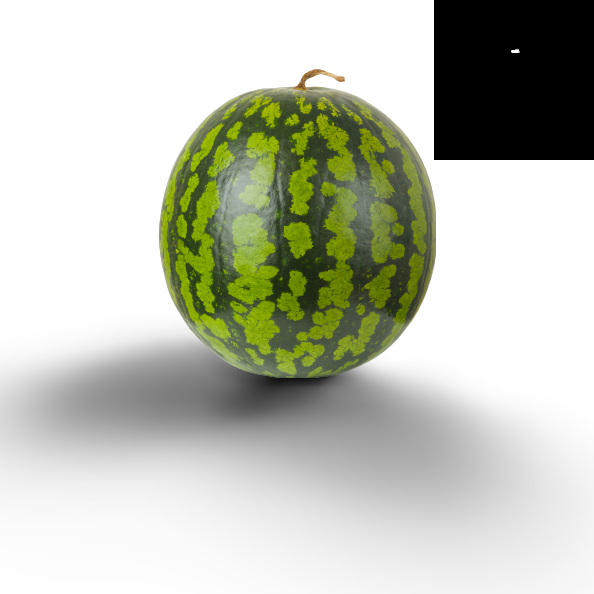}
\caption{An example of the AO refinement. The AO map for SR module is shown in the corner of the composition results. Although AOP predicts a wrong AO map due to the ambiguity of the mask input (left), our SSN can render a much more realistic soft shadow with simple AO refinement (right).} \label{fig:ao_input}
\end{figure}

\subsection{Discussion} 
With the impressive results from recent 2D-to-3D object reconstruction methods, \eg~PIFuHD~\cite{saito2020pifuhd}, one may argue that rendering soft shadows can be straightforward by first getting the 3D object model from an image and then using the traditional shadow rendering methods in computer graphics. However, in general, a 3D object reconstruction task from a single image is still very challenging. Moreover, existing methods like PIFuHD are trained in the natural image domain. Thus it can be difficult for them to generalize to other image domains such as cartoons and paintings.

Besides, there is an even more critical issue for soft shadow rendering using the 2D-to-3D reconstruction methods. In Fig.~\ref{fig:comparison_pifuhd}, we show an example using PIFuHD and the Mitsuba~\cite{jakob2010mitsuba} renderer to generate the soft shadow for a 2D person image. Due to the inaccuracy in the 3D shape and the simplified camera assumption in PIFuHD, the feet' pose may not sufficiently align with the ground plane, making it less controllable to generate desirable shadow effects near the contact points. In contrast, our method can cause different shadow effects near the contact points by modifying the AO map, which provides more control over the 3D geometry interpretation near the contact points. Another example is shown in Fig.~\ref{fig:ao_input}. The mask input for the watermelon is almost a disk which is very ambiguous. With a simple refinement of the AO input, the 3D relationship between the cutout object and the ground is visually more reasonable from the hint of a more realistic soft shadow.  

\paragraph{Limitations} The input to our SSN is an object mask that is a general representation of the 2D object shape. Still, it can be ambiguous for inferring the 3D shape and pose for some objects. Some of the ambiguous cases can be improved by refining the AO map, while others may not be easily resolved in our framework. Some examples are shown in the supplementary materials. Also, the shape of soft shadows depends on camera parameters. However, the mask input is ambiguous for some extreme camera settings. For instance, a very large field-of-view distorts the shadows that the SSN cannot handle. Moreover, we assume our objects are always standing on a ground plane, and the SSN cannot handle the cases where objects are floating above the ground or the shadow receiver is more complicated than a ground plane.



%% file: src/7_conclusion.tex
\section{Conclusion}
We introduced Soft Shadow Network(SSN) to synthesize soft shadows given 2D masks and environment map configurations for image compositing. Naively generating diverse soft shadow training data is cost expensive. To address this issue, we constructed a set of soft shadow bases combined with fast ELM sampling which allowed for fast training and better generalization ability. We also proposed the inverse shadow domain that has significantly improved the convergence rate and overall performance. In addition, a controllable pipeline is proposed to alleviate the generalization limitation by introducing a modifiable ambient occlusion map as input. Experiments on the benchmark demonstrated the effectiveness of our method. User studies confirmed the visual quality and showed that the user can quickly and intuitively generate soft shadows even without any computer graphics experience.




%% file: src/8_acknowledgement.tex
\section*{Acknowledgment}
We appreciate constructive comments from reviewers. Thank Lu Ling for help in data analysis. This research was funded in part by National Science Foundation grant \#10001387, \textit{Functional Proceduralization of 3D Geometric Models}. This work was supported partly by the Adobe Gift Funding. 